\documentclass[conference]{IEEEtran}
\IEEEoverridecommandlockouts
\usepackage{booktabs} 
\usepackage{graphicx}
\usepackage{setspace}
\usepackage{amsmath}
\usepackage{dsfont}
\usepackage{amsfonts}
\usepackage{lipsum}
\usepackage{multicol}
\usepackage{float}
\usepackage{bm}
\usepackage{color}
\usepackage[dvipsnames]{xcolor}
\usepackage{etoolbox}
\usepackage{soul}
\usepackage{amssymb}
\usepackage[linesnumbered,ruled,vlined]{algorithm2e}
\usepackage{mathtools,amssymb}
\usepackage{caption}
\usepackage{float}
\usepackage{amsthm}
\usepackage[caption = false,subrefformat=parens,labelformat=parens]{subfig}
\usepackage{adjustbox}
\usepackage{afterpage}
\usepackage{multirow}
\usepackage{mathrsfs}
\usepackage{hyperref}
\usepackage{tabularx}
\usepackage{array}

\def\BibTeX{{\rm B\kern-.05em{\sc i\kern-.025em b}\kern-.08em
    T\kern-.1667em\lower.7ex\hbox{E}\kern-.125emX}}
\begin{document}

\title{Improving Generalizability of Kolmogorov–Arnold Networks via Error-Correcting Output Codes

\author{
Youngjoon Lee\textsuperscript{\rm 1}, Jinu Gong\textsuperscript{\rm 2}, Joonhyuk Kang\textsuperscript{\rm 1} \\
\textsuperscript{\rm 1}School of Electrical Engineering, KAIST, South Korea\\
\textsuperscript{\rm 2}Department of Applied AI, Hansung University, South Korea\\
Email: yjlee22@kaist.ac.kr, jinugong@hansung.kr, jkang@kaist.ac.kr
}

\thanks{This research was supported by the Institute of Information \& Communications Technology Planning \& Evaluation (IITP)-ITRC (Information Technology Research Center) grant funded by the Korea government (MSIT) (IITP-2025-RS-2020-II201787).\\
}
}

\maketitle

\begin{abstract}
Kolmogorov–Arnold Networks (KAN) offer universal function approximation using univariate spline compositions without nonlinear activations.  
In this work, we integrate Error-Correcting Output Codes (ECOC) into the KAN framework to transform multi‐class classification into multiple binary tasks, improving robustness via Hamming distance decoding.  
Our proposed KAN with ECOC framework outperforms vanilla KAN on a challenging blood cell classification dataset, achieving higher accuracy across diverse hyperparameter settings.  
Ablation studies further confirm that ECOC consistently enhances performance across FastKAN and FasterKAN variants.  
These results demonstrate that ECOC integration significantly boosts KAN generalizability in critical healthcare AI applications.
To the best of our knowledge, this is the first work of ECOC with KAN for enhancing multi-class medical image classification performance.
\end{abstract}


\section{Introduction}
In recent years, deep learning has transformed healthcare through models that analyze medical data with unprecedented accuracy \cite{bisio2025ai, baker2023artificial}.
Neural architectures from MLP \cite{rumelhart1986learning} to Transformers \cite{vaswani2017attention} have revolutionized medical classification by employing their unique mathematical principles to effectively capture patterns in diverse healthcare datasets \cite{rajpurkar2022ai}.
These AI models have advanced automated diagnosis in imaging and pathology, enhancing clinical decision \cite{rajpurkar2023current}. 
Despite their effectiveness, traditional neural networks lack transparency, hindering adoption in healthcare where decision interpretability is crucial \cite{yang2025medkan}. 
Therefore, medical applications represent an ideal domain for interpretable architectures that maintain performance while providing clear mathematical reasoning.

Kolmogorov–Arnold Networks (KAN) \cite{liu2024kan1, liu2024kan2} have emerged as a promising alternative to MLP for deep learning tasks by leveraging the Kolmogorov-Arnold representation theorem \cite{kolmogorov1961representation}. 
Unlike conventional neural networks, KAN utilize univariate spline functions composed in a hierarchical structure to achieve universal function approximation without relying on nonlinear activation functions \cite{lee2025unified}. 
Furthermore, these networks offer enhanced interpretability through their explicit mathematical formulation while maintaining competitive performance across various tasks \cite{somvanshi2024survey}. 
However, despite these advantages, KAN faces challenges in generalizability when applied to complex multi-class classification problems. 
Consequently, current KAN implementations often require meticulous hyperparameter tuning, which can lead to reduced generalization capability in real-world scenarios.

\begin{figure}[t]
    \centering
    \includegraphics[width=\linewidth]{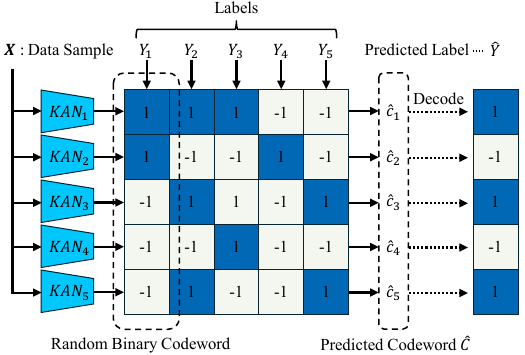}
    \caption{Illustration of the proposed KAN with ECOC framework. The input data sample $X$ is processed through multiple KAN classifiers ($KAN_1$ to $KAN_5$), each trained to predict a specific bit position in the binary codeword. The binary outputs from each KAN form the predicted codeword $\hat{C}$, which is then decoded to the nearest valid class codeword based on Hamming distance, resulting in the final predicted label $\hat{Y}$.}
    \label{fig:1}
\end{figure}

Error-Correcting Output Codes (ECOC) represent an established ensemble method that transforms multi-class classification problems into multiple binary classification tasks \cite{dietterich1994solving, kong1995error}. 
By encoding class labels as binary codewords with sufficient Hamming distance between classes, ECOC provides redundancy that enhances robustness against classification errors \cite{deng2010applying}. 
Specifically, this approach has successfully improved generalization across various classifiers including decision trees and neural networks \cite{yu2024error}. 
Moreover, ECOC leverages the collective predictions of multiple binary classifiers, allowing for error correction during the decoding phase when predictions are mapped back to the original class space \cite{verma2019error}. 
Thus, the integration of ECOC with modern neural architectures enhances model robustness, making it particularly valuable for applications requiring high reliability.

Illustration of our proposed KAN with ECOC framework is presented in Fig.~\ref{fig:1}, demonstrating how we combine the structural advantages of KAN with the error-correcting properties of ECOC. 
The framework transforms the original multi-class problem into a set of binary classification tasks, where each binary classifier corresponds to a bit position in the codeword matrix. 
During training, KAN learns to predict individual bits of the codeword rather than directly predicting class labels. 
At inference time, the binary outputs from all KAN form a predicted codeword which is decoded to the nearest valid class codeword according to Hamming distance. 
Additionally, this approach provides robustness against individual classifier errors while maintaining the interpretability benefits of KAN.

The main contributions of this paper are as follows:
\begin{itemize}
    \item We propose a novel integration of KAN with ECOC to enhance performance in multi-class classification tasks.
    \item We demonstrate that KAN with ECOC outperforms vanilla KAN across blood cell classification task.
    \item We show that the performance improvements are consistent across different hyperparameter configurations.
    \item Additionally, ablation studies demonstrate that ECOC improves performance across KAN variants—including FastKAN \cite{li2024kolmogorov} and FasterKAN \cite{Athanasios2024}.
\end{itemize}

The remainder of this paper is organized as follows.  
In Section~\ref{sec:method}, we describe our integration of ECOC into KAN for multi‐class classification.  
Section~\ref{sec:experiment} presents quantitative results on blood cell classification and ablation studies across FastKAN and FasterKAN.  
Finally, Section~\ref{sec:conclusion} concludes the paper and outlines future research directions.

\section{System Model}
\label{sec:method}

In this section, we present KAN for universal function approximation and ECOC for error-correcting multi-class decomposition. 
Our framework integrates these approaches to enhance general classification performance and robustness in challenging medical imaging applications.

\subsection{Kolmogorov-Arnold Networks}
KAN implements the Kolmogorov-Arnold representation theorem, which states that any multivariate continuous function $f: [0,1]^n \rightarrow \mathbb{R}$ can be expressed as:

\begin{equation}
f(x_1, x_2, \ldots, x_n) = \sum_{q=1}^{2n+1}\Phi_q\left(\sum_{p=1}^{n}\phi_{q,p}(x_p)\right),
\end{equation}
where $\phi_{q,p}$ and $\Phi_q$ are continuous univariate functions.

Building upon this theoretical foundation, KAN utilizes univariate spline functions as fundamental building blocks, whereby each layer performs the operation:
\begin{equation}
z_j^{(l)} = \sum_{i=1}^{d_{l-1}} g_{ij}^{(l)}(z_i^{(l-1)}),
\end{equation}
where $z_j^{(l)}$ is the $j$-th output of layer $l$, $d_{l-1}$ is the previous layer's dimensionality, and $g_{ij}^{(l)}$ are learnable univariate functions implemented as B-spline curves.

For multi-class problems with $k$ classes, KAN's final layer transforms the network outputs into logits $\{o_1, o_2, \ldots, o_k\}$ which are subsequently converted into class probabilities using the softmax function:
\begin{equation}
p(y=c|x) = \frac{\exp(o_c)}{\sum_{j=1}^{k}\exp(o_j)}.
\end{equation}

The KAN training proceeds by minimizing the cross-entropy loss function:
\begin{equation}
\mathcal{L}_{\text{CE}} = -\sum_{i=1}^{N}\sum_{c=1}^{k}y_{i,c}\log(p(y=c|x_i)).
\end{equation}

\subsection{Error-Correcting Output Codes}
The ECOC transforms multi-class classification into multiple binary classification tasks by encoding class labels as binary codewords with sufficient Hamming distance between classes. 
In detail, for a $k$-class problem, ECOC defines a coding matrix $M \in \{-1, 1\}^{k \times b}$, where $b$ is the codeword length, typically set to $b = 2k$ to ensure sufficient code separation for optimal error correction. 
The ECOC consists of two principal components:
\begin{itemize}
    \item \textbf{Encoding}: Each class $i$ is assigned a unique binary codeword $M_i = \{c_{i1}, c_{i2}, \ldots, c_{ib}\}$ where $c_{ij} \in \{-1, 1\}$. The coding matrix is generated stochastically according to:
    \begin{equation}
    M_{i,j} \sim 2 \cdot \text{Bernoulli}(0.5) - 1.
    \end{equation}
    This random coding strategy ensures that the expected Hamming distance between any two codewords is approximately $b/2 = k$, thereby providing robust error correction capabilities.
    
    \item \textbf{Decoding}: For prediction, the predicted codeword $\hat{C} = \{\hat{c}_1, \hat{c}_2, \ldots, \hat{c}_b\}$ is compared to each row in the coding matrix, and the class with the minimum Hamming distance is selected:
    \begin{equation}
    \hat{y} = \arg\min_{i \in \{1,\ldots,k\}} d_H(M_i, \hat{C}),
    \end{equation}
    where the Hamming distance between two binary codewords is defined as $d_H(M_i, \hat{C}) = \sum_{j=1}^{b} \mathds{1}(c_{ij} \neq \hat{c}_j).$
\end{itemize}
Subsequently, for binary classification tasks within the ECOC method, we employ the binary cross-entropy loss function:
\begin{equation}
\mathcal{L}_{\text{BCE}} = -\frac{1}{N}\sum_{i=1}^{N}\left[y_i\log(\sigma(o_i)) + (1-y_i)\log(1-\sigma(o_i))\right],
\end{equation}
where $\sigma(o_i) = \frac{1}{1 + e^{-o_i}}$ is the sigmoid function.

\begin{figure*}[t]
  \centering
  \includegraphics[width=\textwidth]{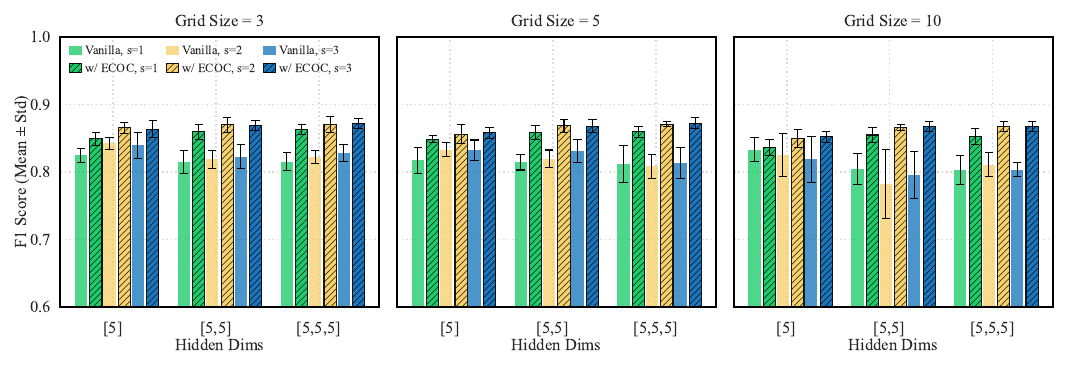}
  \caption{F1 scores (mean $\pm$ std) of vanilla KAN versus KAN with ECOC across different hyperparameter configurations. Results show performance for varying grid sizes (3, 5, 10), spline orders (s=1, 2, 3), and hidden layer dimensions ([5], [5,5], [5,5,5]). Striped bars represent our proposed KAN with ECOC and solid bars denote vanilla KAN performance.}
  \label{fig:fig2}
  \vspace{-0.3cm}
\end{figure*}

\subsection{Integration Framework}
Our framework integrates KAN with ECOC to address generalizability challenges in medical multi-class classification task. 
The integration methodically decomposes the original multi-class problem into binary classification tasks, trains separate KAN models for each binary task, and combines their outputs during inference.

As illustrated in Fig.~\ref{fig:1}, the framework consists of the following components:
\begin{enumerate}
    \item \textbf{Coding Matrix Construction}: The proposed method uses the previously defined coding strategy to generate matrix $M \in \{-1, 1\}^{k \times 2k}$.
    
    \item \textbf{Binary Classifier Training}: For each bit position $j \in \{1, 2, \ldots, b\}$ in the codeword, we train a corresponding KAN classifier $KAN_j$ to predict whether the input sample belongs to a class with $c_{ij} = 1$ or $c_{ij} = -1$. Each binary classifier is trained using the binary cross-entropy loss $\mathcal{L}_{\text{BCE}}$ defined earlier.
    
    \item \textbf{Prediction Aggregation}: During inference, for a given input sample $x$, each binary KAN classifier $KAN_j$ produces a binary prediction $\hat{c}_j \in \{-1, 1\}$. These individual predictions are aggregated to form the predicted codeword $\hat{C} = \{\hat{c}_1, \hat{c}_2, \ldots, \hat{c}_b\}$.
    
    \item \textbf{Hamming Distance Decoding}: The final class prediction is determined using the Hamming distance decoding approach defined in the ECOC section.
\end{enumerate}
This integration enhances robustness through ECOC error-correction, and reduces hyperparameter sensitivity, collectively improving model generalization.

\section{Experiment and Results}\label{sec:experiment}
\subsection{Experiment Setting}
We evaluated our KAN with ECOC framework using the blood cell classification dataset \cite{acevedo2020dataset}, which contains microscopic images of eight different blood cell types. 
Experiments were conducted across different hyperparameter settings, including spline orders ($s$), grid sizes, and hidden layer configurations. 
All models were trained for 100 epochs using Adam optimizer with a learning rate of $0.001$, with experiments repeated across six different random seeds.
Note that, we utilized EfficientKAN for computational efficiency.
Our implementation is available in our open source repository\footnote{https://github.com/yjlee22/kan-ecoc}.

\subsection{Results}

\begin{table}[h]
\centering
\caption{Performance comparison of vanilla KAN and our proposed KAN with ECOC across multiple metrics.}
\begin{tabularx}{\columnwidth}{>{\raggedright\arraybackslash}X>{\centering\arraybackslash}X>{\centering\arraybackslash}X}
\toprule
\textbf{Metric} & \textbf{Vanilla KAN} & \textbf{KAN w/ ECOC} \\
\midrule
Accuracy & $0.8309\pm0.0154$ & $0.8519\pm0.0117$ \\
Recall & $0.8309\pm0.0154$ & $0.8519\pm0.0117$ \\
Precision & $0.8372\pm0.0147$ & $0.8690\pm0.0064$ \\
F1 Score & $0.8320\pm0.0154$ & $0.8581\pm0.0081$ \\
\bottomrule
\end{tabularx}
\label{tab:0}
\end{table}

\subsubsection{Impact of ECOC}
To investigate the impact of ECOC integration on KAN performance, we analyze the results in Table \ref{tab:0} comparing vanilla KAN with our proposed approach. 
Our KAN-ECOC framework consistently outperforms the vanilla KAN across all metrics, demonstrating a clear performance advantage. 
The accuracy and recall show identical improvements from 0.8309 to 0.8519, while precision exhibits the most substantial gain from 0.8372 to 0.8690. 
Furthermore, the F1 score increases by 0.0261 points from 0.8320 to 0.8581, confirming the overall effectiveness of our approach. 
The results indicate that ECOC integration strengthens KAN with improved generalization.

\subsubsection{Impact of Hyperparameter Configuration}
To investigate how hyperparameter variations influence model performance, we conducted experiments across diverse configurations.
As shown in Fig. \ref{fig:fig2}, the performance gap is maintained regardless of grid size (3, 5, or 10), with the most significant improvements appearing in deeper networks ($[5,5]$ and $[5,5,5]$). 
While both methods benefit from increased spline complexity ($s=2$, $s=3$), our ECOC integration provides additional gains at every spline order level. 
Notably, ECOC's performance advantage remains consistent across all grid resolutions, demonstrating that our approach enhances generalization independently of this parameter. 
Thus, KAN-ECOC provides more stable performance across varying hyperparameter settings.

\begin{figure}[h]
  \centering
  \includegraphics[width=\columnwidth]{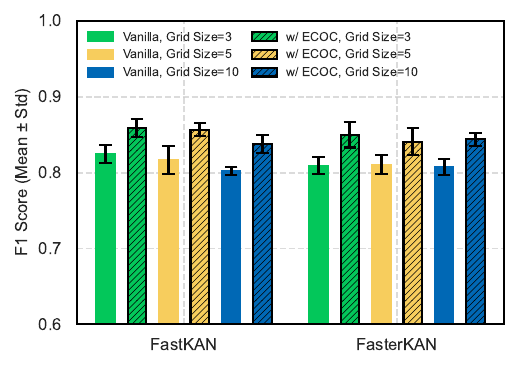}
  \caption{Comparison of F1 scores (mean $\pm$ std) with vanilla and ECOC-integrated versions of FastKAN and FasterKAN across different grid sizes.}
  \label{fig:fig3}
\end{figure}

\subsubsection{Ablation Study}
To verify the effectiveness of our approach across different KAN variants, we conducted an ablation study using FastKAN and FasterKAN as shown in Fig. \ref{fig:fig3}.
These variants are chosen to evaluate whether ECOC consistently enhances performance across both lightweight and optimized versions of KAN.
Our results confirm that ECOC integration consistently improves performance across both architectures and all grid sizes (3, 5, 10), demonstrating the robustness of our method. 
Despite FasterKAN showing slightly lower baseline performance than FastKAN, both variants exhibit similar relative improvements with ECOC integration, indicating our approach provides benefits independent of specific KAN implementations. 
This consistent enhancement across different architectures further validates that ECOC is an effective technique for improving KAN generalization regardless of model configuration.

\section{Conclusion}\label{sec:conclusion}
In this work, we introduce a novel framework integrating KAN with ECOC to enhance multi-class classification performance in medical imaging applications. 
The performance gains remain stable across various hyperparameter settings, demonstrating our framework's robustness to grid sizes, spline orders, and network architectures. 
Additionally, ablation studies with FastKAN and FasterKAN variants confirm that ECOC integration provides consistent benefits regardless of the KAN implementation. 
As future work, we plan to extend KAN–ECOC to federated learning, enabling privacy-preserving solutions in distributed healthcare settings.

\bibliographystyle{IEEEtran}
\bibliography{reference}

\end{document}